\documentclass[letterpaper]{article} 
\usepackage{aaai25}  
\usepackage{times}  
\usepackage{helvet}  
\usepackage{courier}  
\usepackage[hyphens]{url}  
\usepackage{graphicx} 
\def\UrlFont{\rm}  
\usepackage{natbib}  
\usepackage{caption} 
\usepackage{amsmath}
\usepackage{amssymb} 
\usepackage{multirow}
\usepackage{algorithm}
\usepackage{algorithmic}
\newenvironment{new_boxed}
    {\begin{center}
    \begin{tabular}{|p{0.9\linewidth}|}  
    \hline\\
    }
    { 
    \\\\\hline
    \end{tabular} 
    \end{center}
    }

\usepackage{newfloat}
\usepackage{listings}
\DeclareCaptionStyle{ruled}{labelfont=normalfont,labelsep=colon,strut=off} 
\floatstyle{ruled}
\newfloat{listing}{tb}{lst}{}
\floatname{listing}{Listing}
\title{DELIA: \\Diversity-Enhanced Learning for Instruction Adaptation in Large Language Models}
\author{
    Yuanhao Zeng\textsuperscript{\rm 1},
    Fei Ren\textsuperscript{\rm 1},
    Xinpeng Zhou\textsuperscript{\rm 2},
    Yihang Wang\textsuperscript{\rm 1},
    Yingxia Shao\textsuperscript{\rm 1}
}
\affiliations{
    \textsuperscript{\rm 1}Beijing University of Posts and Telecommunications, \\
    \textsuperscript{\rm 2}Harbin Institute of Technology, \\
    \{cengyuanhao, renfei\}@bupt.edu.cn\\
    2021113676@stu.hit.edu.cn
}

\usepackage{bibentry}

\begin{document}

\maketitle

\begin{abstract}
Although instruction tuning is widely used to adjust behavior in Large Language Models (LLMs), extensive empirical evidence and research indicates that it is primarily a process where the model fits to specific task formats, rather than acquiring new knowledge or capabilities. We propose that this \textbf{limitation stems from biased features learned during instruction tuning}, which differ from ideal task-specfic features, leading to learn less underlying semantics in downstream tasks. However, ideal features are unknown and incalculable, constraining past work to rely on prior knowledge to assist reasoning or training, which limits LLMs' capabilities to the developers' abilities, rather than data-driven scalable learning. In our paper, through our novel data synthesis method, \textbf{DELIA (Diversity-Enhanced Learning for Instruction Adaptation)}, we leverage the buffering effect of extensive diverse data in LLMs training to transform biased features in instruction tuning into \textbf{approximations of ideal features, without explicit prior ideal features}. Experiments show DELIA's better performance compared to common instruction tuning and other baselines. It outperforms common instruction tuning by \textbf{17.07\%-33.41\%} on Icelandic-English translation bleurt score (WMT-21 dataset, gemma-7b-it) and improves accuracy by \textbf{36.1\%} on formatted text generation (Llama2-7b-chat). Notably, among knowledge injection methods we've known, DELIA uniquely align the internal representations of new special tokens with their prior semantics. 
\end{abstract}

%

\section{Introduction}

Large Language Models (LLMs) have demonstrated remarkable capabilities across a wide range of tasks. However, their application in specific domains often requires additional fine-tuning to align with particular use cases. Instruction tuning has emerged as a popular method to address this limitation, aiming to endow LLMs with the ability to follow task-specific instructions.

Instruction tuning, typically involving supervised fine-tuning on instruction-response pairs, has been shown to primarily fit models to specific task formats rather than imparting new knowledge or capabilities \citep{Zhao2024-mm, Zhou2023-ag}. This limitation is particularly evident with smaller datasets \citep{Zhou2023-ag}, contradicting the ideal scenario where LLMs learn adaptable downstream task capabilities.

We propose that this issue arises from the discrepancy between instruction tuning data distribution and the diverse real-world instruction distribution, leading to the learning of biased features during instruction tuning. These biased features deviate from ideal task-specific features. While not explicitly framed as such, previous works have implicitly acknowledged this feature bias in instruction tuning datasets as a key factor in suboptimal performance \citep{Zhao2024-mm, Patel2024-jf}.

While ideal features may be unknown and incalculable, approximating them is crucial for grasping task semantics. Previous works have attempted to address this challenge from the perspectives of model architecture and training objectives\citep{Wang2022-oh, Vernikos2020-hn}. However, given the robust library and hardware support for LLMs, it is crucial to bridge the gap between biased and ideal features without altering the model's structure and at a low cost. 

Synthetic data offers a promising solution, but past methods often relied on heuristic approaches based on developers' prior knowledge \citep{Li2023-vk, Zhao2024-mm}. To transcend developer limitations, we believe it is necessary to extend model capabilities through data-driven methods. The question then becomes: how can data-driven synthetic data methods approximate ideal features?

To address this challenge, we introduce DELIA (Diversity-Enhanced Learning for Instruction Adaptation). We demonstrate that sufficiently diverse data, capable of covering as many potential features as possible, can act as a buffer, mitigating the bias of biased features relative to ideal features and approximating ideal features.

DELIA leverages the buffering effect of extensive diverse data in LLMs training to transform biased features in instruction tuning into approximations of ideal features, without explicit training objectives towards ideal features. Our method involves sampling diverse question-answer pairs from LLMs, anisotropically diversifying downstream task instructions, and extensively shuffling these components for training.

Our empirical evaluation of DELIA demonstrates significant improvements over common instruction tuning methods and other baselines across various tasks. Furthermore, among knowledge injection methods we've known, DELIA uniquely align the internal representations of new special tokens with their prior semantics, which previous researches generally considered difficult to achieve \citep{Wang2022-dl, sun2024amurocharanalyzing}.

\begin{figure*}[t]
\centering
\includegraphics[width=0.9\textwidth]{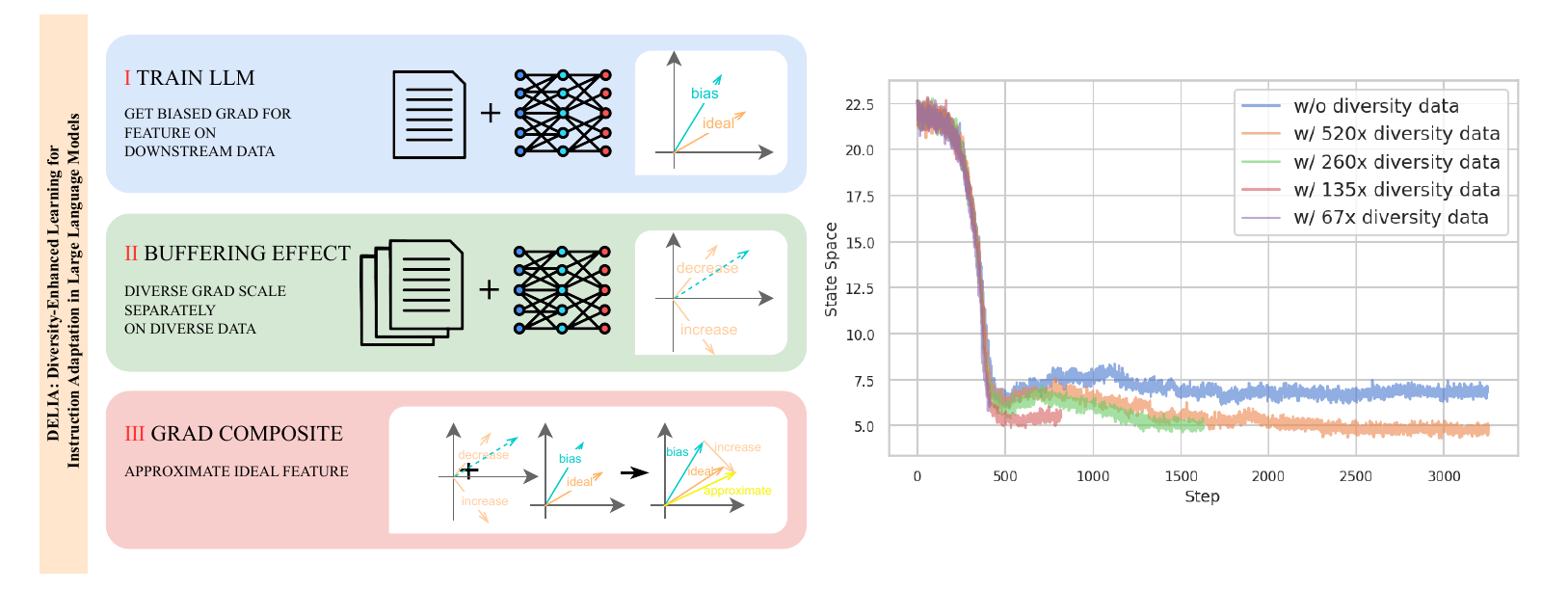}
\caption{overview of DELIA and buffering effect}
\label{fig1}
\end{figure*}

The contributions of our work can be summarized as follows:

\begin{itemize}
    \item We propose a novel modeling approach that attributes the limitations of instruction tuning in learning downstream task capabilities to biased features in datasets. This perspective provides a new understanding of the challenges in instruction tuning.
    \item We demonstrate the existence of a buffering effect and, based on this, introduce our DELIA (Diversity-Enhanced Learning for Instruction Adaptation) method. DELIA utilizes the buffering effect of extensive diverse data in LLMs training to approximate ideal features. Our empirical results show that DELIA significantly outperforms common instruction tuning methods and other baselines across various tasks.
    \item We provide evidence that DELIA enables LLMs to align the internal representations of new special tokens with their prior semantics, surpassing known knowledge injection methods. This capability represents a significant advancement in the field of language model adaptation and knowledge integration.
\end{itemize}


\section{Related Work}

\subsection{Limitations of Instruction Tuning}

Instruction tuning has become a widely adopted method for enhancing model performance on specific tasks, but numerous studies have revealed inherent flaws in this approach. \citet{Sclar2023-bu} demonstrated that such models could become excessively sensitive to prompt variations, compromising their robustness. \citet{sun2024amurocharanalyzing} observed that fine-tuning often results in models fitting to specific task formats, hindering their ability to adapt to novel presentations of similar tasks. \citet{Wang2022-dl} argued that instruction-tuned models become overly specialized in specific task formats rather than learning underlying semantics, which 
is basically near what we proposed ``biased feature".

\subsection{Approaches to Improving Instruction Following}

Researchers have proposed various solutions to address the challenges of instruction tuning. At the framework level,\citet{Li2023-vk} and \citet{Zhao2024-mm} proposed frameworks that iteratively self-improves training data quality. Addressing the problem at the training objective level, \citet{Vernikos2020-hn} proposed modifying the training objective with an adversarial classifier to mitigate domain overfitting. Architectural modifications have also been explored. \citet{Wang2022-oh} explored architectural modifications with multiple PEFT modules. \citet{Jain2023-em} focused on altering the training process itself, by introducing noise during embedding to improve generalization capabilities.

\subsection{Synthetic Data for Model Improvement}

Synthetic data generation has emerged as a promising solution to instruction tuning challenges, offering the introduction of human-interpretable prior knowledge and scalable, data-driven performance improvement.\citet{Liu2023-pt} explored manipulating instruction positioning in training data. \citet{dong-etal-2024-abilities} focuses on balancing specialized and general knowledge retention. \citet{Mecklenburg2024-uj} involves extracting and synthesizing atomic facts for more robust knowledge representation. These approaches demonstrate the potential of synthetic data in addressing the limitations of traditional instruction tuning methods, providing a foundation for our work in approximating ideal task features and improving model performance across various downstream tasks.

\section{Method}

\subsection{Problem Definition}

The training objective of Large Language Models (LLMs) is to minimize the cross-entropy loss function. In the context of instruction tuning, for a given instruction, we aim to minimize the following loss:

\begin{equation}
    \mathcal{L}(\theta) = -\mathbb{E}_{x \sim p(x)}[\sum_{t=ins}^{response} \log \hat{q}_\theta(x_t | x_{<t})]
\end{equation}

where \( p \) is the true distribution and \(\hat{q}\) is the predicted distribution.

However, there is a discrepancy between the ideal distribution of downstream tasks \(p_{d}\) and the general alignment distribution \(p_{g}\) of open-source LLMs. Our goal is to enable LLMs to acquire downstream task capabilities through instruction tuning, i.e., to find parameters \(\theta_{d}\) that minimize:

\begin{equation}
    \mathcal{L}(\theta_d) = -\mathbb{E}_{x \sim p_d(x)}[\sum_{t=ins}^{response} \log \hat{q}_{\theta_d}(x_t | x_{<t})]
\end{equation}

This leads to a difference between the model parameters \(\theta_{d}\) adapted for downstream tasks and the aligned parameters \(\theta_{g}\).

\subsubsection{Ideal Features vs. Biased Features}

We expect LLMs to learn ideal features for downstream tasks through $p_d$, enabling task completion under any instructions with similar semantics. Based on this, training objectives corresponding to $p_d$ and $p_g$ should have similar loss curve trends.

However, the ideal distribution \(p_{d}\) for downstream tasks is unknown and incalculable. We can only approximate $p_d$ using the distribution $p_d'$ of the instruction tuning dataset. This introduces several sources of bias, including bias from instruction phrasing and overfitting bias due to fixed training instructions.

These biases produce biased features, causing the actual training distribution $p_d'$ to deviate from the ideal downstream task distribution \(p_{d}\). Consequently, the actual loss function becomes:

\begin{equation}
    \mathcal{L}(\theta_d') = -\mathbb{E}_{x \sim p_d'(x)}[\sum_{t=ins}^{response} \log \hat{q}_{\theta_d'}(x_t | x_{<t})]
\end{equation}

Practice shows that using this actual loss function as the optimization objective can achieve meaningful fit on downstream tasks with vast data or extensive training. However, this is primarily a process of fitting specific task formats rather than acquiring new knowledge or capabilities. We believe this is due to the difference between biased features corresponding to $p_d'$ and ideal features corresponding to $p_d$. To enable LLMs to ideally complete tasks under any instructions with similar semantics, we need to bridge the gap between biased and ideal features.

\subsubsection{Distribution Difference Hypothesis}

Based on empirical observations, we hypothesize that $p_{d}$ is closer to $p_{g}$ than $p_{d}'$, i.e.:

\begin{equation}
    D_{kl}(p_{d}'||p_{g}) > D_{kl}(p_{d}||p_{g})
\end{equation}

This divergence between $p_{d}'$ and $p_{d}$ arises from our goal of enabling LLMs to perform downstream tasks under any instructions with similar semantics after instruction tuning. However, the training data inevitably contains biases, such as bias in instruction wording or overfitting bias due to fixed training instructions as discussed.

These biases lead to the biased features in $p_{d}'$. Importantly, these specific biases are relatively rare in the general alignment distribution $p_{g}$, which explains our hypothesis that $p_{d}$ is closer to $p_{g}$ than $p_{d}'$.

\subsection{DELIA Method: Bridging the Gap Between Biased and Ideal Features}

To address the discrepancy in features, we propose the DELIA (Diversity-Enhanced Learning for Instruction Adaptation) method. The core idea is to introduce extensive diverse data to mitigate the gap between $p_d'$ and $p_d$.

\subsubsection{Buffering Effect}

We posit that introducing extensive diverse training data sampled from similar LLMs can produce the following effects:

\begin{enumerate}
    \item When trained independently, these data generate gradients that largely cancel each other out, resulting in minimal changes to the LLM's parameters.
    \item When trained together with downstream task data, the training process:
    \begin{itemize}
        \item Reduces the gradients produced by data similar to the downstream task
        \item Increases the gradients produced by data significantly different from the downstream task
    \end{itemize}
\end{enumerate}

This gradient differential allows the model to learn biased features while also learning content biased towards general features. We term this the buffering effect, as shown in Figure \ref{fig1}.

\subsubsection{Mathematical Derivation}

To more precisely illustrate the working principle of the DELIA method, we provide the following mathematical derivation:

Let $p_d'(x)$, $p_g(x)$, and $q_\theta(x)$ denote the probability density functions of the downstream task dataset, the extensive diverse dataset, and the model prediction, respectively. Let $\theta$ be the model parameters and $\mathcal{L}(\theta, p)$ be the cross-entropy loss function when training on a dataset with distribution $p$.

Since the LLM has been trained on a subset of the downstream task dataset, we define $\epsilon(x)$ such that:

\begin{equation}
    q_\theta(x) = p_d'(x) + \epsilon(x)
\end{equation}

When training on the extensive diverse dataset, the gradient of the LLM is:

\begin{equation}
    \nabla_\theta \mathcal{L}(\theta, p_g) = \mathbb{E}_{x \sim p_g(x)}[\nabla_\theta (-\log q_\theta(x))]
\end{equation}

Substituting $q_\theta(x)$, we get:

\begin{equation}
    \nabla_\theta \mathcal{L}(\theta, p_g) = \mathbb{E}_{x \sim p_g(x)}[\nabla_\theta (-\log (p_d'(x) + \epsilon(x)))]
\end{equation}

Applying Taylor expansion at $p_d'(x)$ and taking the first-order approximation:

\begin{equation}
    -\log(p_d'(x) + \epsilon(x)) \approx -\log p_d'(x) - \frac{\epsilon(x)}{p_d'(x)}
\end{equation}

Therefore, the gradient can be approximated as:

\begin{align}
\nabla_\theta \mathcal{L}(\theta, p_g) &\approx \mathbb{E}_{x \sim p_g(x)}[\nabla_\theta (-\log p_d'(x) - \frac{\epsilon(x)}{p_d'(x)})] \notag \\
&= \mathbb{E}_{x \sim p_g(x)}[\nabla_\theta (-\log p_d'(x))] \notag \\
&\quad - \mathbb{E}_{x \sim p_g(x)}[\nabla_\theta (\frac{\epsilon(x)}{p_d'(x)})] \notag \\
&= \nabla_\theta [H(p_g, p_d')] - \nabla_\theta [\mathbb{E}_{x \sim p_g(x)}(\frac{\epsilon(x)}{p_d'(x)})] \notag \\
&= \nabla_\theta [D_{KL}(p_g \| p_d') + H(p_g)] \notag \\
&\quad - \nabla_\theta [\mathbb{E}_{x \sim p_g(x)}(\frac{\epsilon(x)}{p_d'(x)})]
\end{align}

In practice, due to good convergence, $\epsilon(x)$, which represents the difference between the predicted distribution and the downstream task dataset distribution, is usually small. Therefore, its gradient has little impact on the overall result and can be ignored. For a single data point in the extensive diverse dataset, we can observe:

1. When $p_g(x)$ is close to $p_d'(x)$, $D_{KL}(p_g || p_d')$ is small, resulting in a relatively small gradient $\nabla_\theta \mathcal{L}(\theta, p_d)$.

2. When $p_g(x)$ is far from $p_d'(x)$, $D_{KL}(p_g || p_d')$ is large, resulting in a relatively large gradient $\nabla_\theta \mathcal{L}(\theta, p_d)$.

This is consistent with our previously mentioned buffering effect.

Based on our earlier assumption that $p_d$ and $p_g$ correspond to training objectives with similar loss curves, we can infer:

\begin{align}
\|\nabla_\theta \mathcal{L}(\theta, p_d)\| &\approx \|\nabla_\theta \mathcal{L}(\theta, p_g)\| \notag \\
&= \big\|\nabla_\theta [D_{KL}(p_g \| p_d') + H(p_g)] \notag \\
&\quad - \nabla_\theta [\mathbb{E}_{x \sim p_g(x)}(\frac{\epsilon(x)}{p_d'(x)})]\big\| \\
&> 0 \notag
\end{align}

Furthermore, we can express the similarity in gradient vector directions:

\begin{equation}
    \frac{\nabla_\theta \mathcal{L}(\theta, p_d)}{\|\nabla_\theta \mathcal{L}(\theta, p_d)\|} \approx \frac{\nabla_\theta \mathcal{L}(\theta, p_g)}{\|\nabla_\theta \mathcal{L}(\theta, p_g)\|}
\end{equation}

This derivation suggests that introducing extensive diverse data may transform gradients learned towards $p_d'$ into approximations of gradients learned towards $p_d$. Specifically:

The directions of the gradient vectors are approximately the same, indicating that the optimization process moves in similar directions.
The norms of the gradient vectors are approximately equal, suggesting similar magnitudes of optimization steps.

This similarity in gradients (both in direction and magnitude) supports our hypothesis that by introducing diverse data, we can guide the model to learn representations closer to ideal features rather than merely overfitting to the biased downstream task dataset. This mechanism allows the model to maintain adaptability to downstream tasks while learning more general and robust feature representations.

\subsection{Implementation of the DELIA Method}

\begin{algorithm}[tb]
\caption{DELIA (Diversity-Enhanced Learning for Instruction Adaptation) Method}
\label{alg:DELIA}
\textbf{Input}: Pre-trained LLM, Downstream task dataset $D$\\
\textbf{Parameter}: Sampling ratio $r$, Diversity training iterations $k$\\
\textbf{Output}: Fine-tuned LLM for downstream tasks
\begin{algorithmic}[1]
\STATE Initialize fine-tuned model with pre-trained LLM
\STATE Add special tokens and init embedding weight
\STATE $A \gets$ Sample diverse question-answer pairs from LLM
\STATE $D \gets$ Anisotropically diversify instructions of $D$
\FOR{each data point $d$ in $D$}
    \STATE Train LLM on $d$
    \STATE Update model parameters
    \STATE $M \gets$ Sample $r$ examples from $A$
    \FOR{each sample $m$ in $M$}
        \STATE Train LLM on $m$
        \STATE Update model parameters
    \ENDFOR
\ENDFOR
\STATE \textbf{return} Fine-tuned LLM
\end{algorithmic}
\end{algorithm}

The proposed data synthesis strategy, DELIA, is a simple yet effective approach to enable LLMs to learn downstream task capabilities and grasp the inherent semantics of tasks. The key aspects are:
\begin{enumerate}
    \item Diverse data is crucial - it must cover the vast majority of potential features to allow biased features to approximate ideal features.
    \item In addition to sample-level diversity, word-level diversity is introduced by using GPT-4 to diversify the instructions of downstream task data. New special tokens are added to validate DELIA align the internal representations of new special token with its prior semantics.
    \item While extensive diverse dataset training after each downstream task data point is theoretically designed, the proposed strategy of extensively shuffling the data provides a practical approximation.
\end{enumerate}





\section{Experiment}

In this section, we evaluate the effectiveness of DELIA through the following experiments: (1) verifying DELIA's improvement of the model's intermediate representations, to validate DELIA's role in promoting the model's learning of ideal downstream task features; (2) performance on the practical formatted text generation task; (3) performance on the practical English-Icelandic translation task. Finally, we analyze the ablation experiments on DELIA, indicating that DELIA's effect is not simply from the diversified downstream task data or extensive diversified downstream data, but from the synergistic effect of the two, i.e., the approximation of the ideal task-specific features we emphasize. We also show from the perspective of the state space of the divergent token and the overall loss of the downstream tasks that DELIA can continuously improve the performance of the downstream tasks.

\begin{table*}[h]
\centering
\begin{tabular*}{0.9\textwidth}{@{\extracolsep{\fill}}lcccccc}
\hline
\multirow{2}{*}{Method} & \multicolumn{2}{c}{First Block} & \multicolumn{2}{c}{Middle Block} & \multicolumn{2}{c}{Last Block} \\
& ``json" & ``thought" & ``json" & ``thought" & ``json" & ``thought" \\
\hline
Random Init & 382.5 & 382.3 & 388.0 & 388.1 & 203.5 & 202.1 \\
Mean Embedding & 328.1 & 327.9 & 333.8 & 333.9 & 188.2 & 186.9 \\
Common Instruction Tuning & 336.3 & 337.5 & 341.9 & 343.4 & 182.3 & 183.0 \\
DMT & 271.1 & 265.3 & 277.6 & 272.0 & 168.3 & 168.1 \\
Fact Based & 94.5 & 94.2 & 102.4 & 101.9 & 100.5 & 103.1 \\
DELIA & \textbf{3.2} & \textbf{6.2} & \textbf{15.6} & \textbf{14.4} & \textbf{46.1} & \textbf{47.7} \\
\hline
\end{tabular*}
\caption{\label{table:intermedia}L2 norms of \texttt{<sep>} to keywords in different transformer blocks, the lower the better, suggesting that LLM align the internal representations of new special token with its prior semantics}
\end{table*}

\begin{figure}[t]
\centering
\includegraphics[width=0.9\columnwidth]{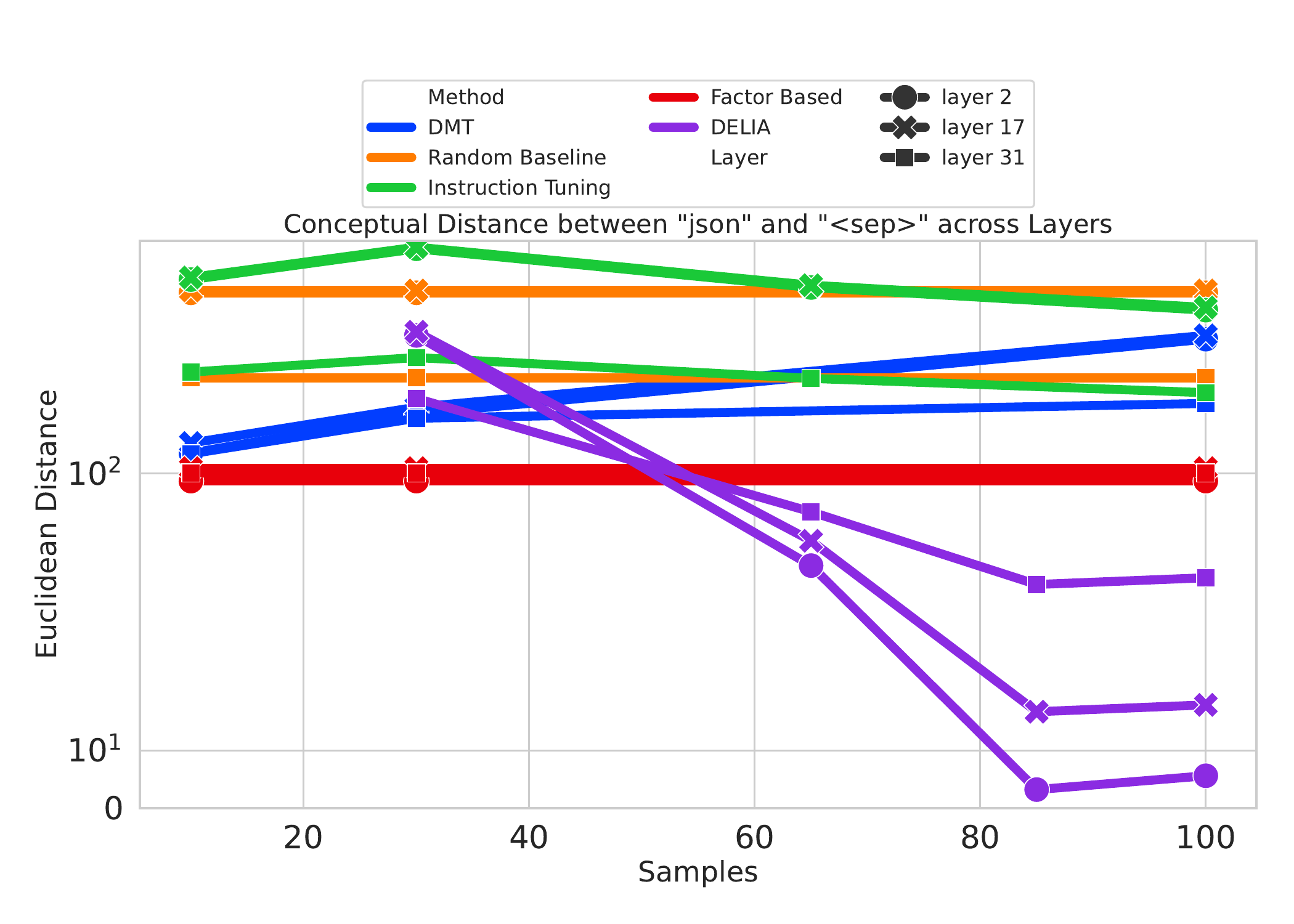}
\includegraphics[width=0.9\columnwidth]{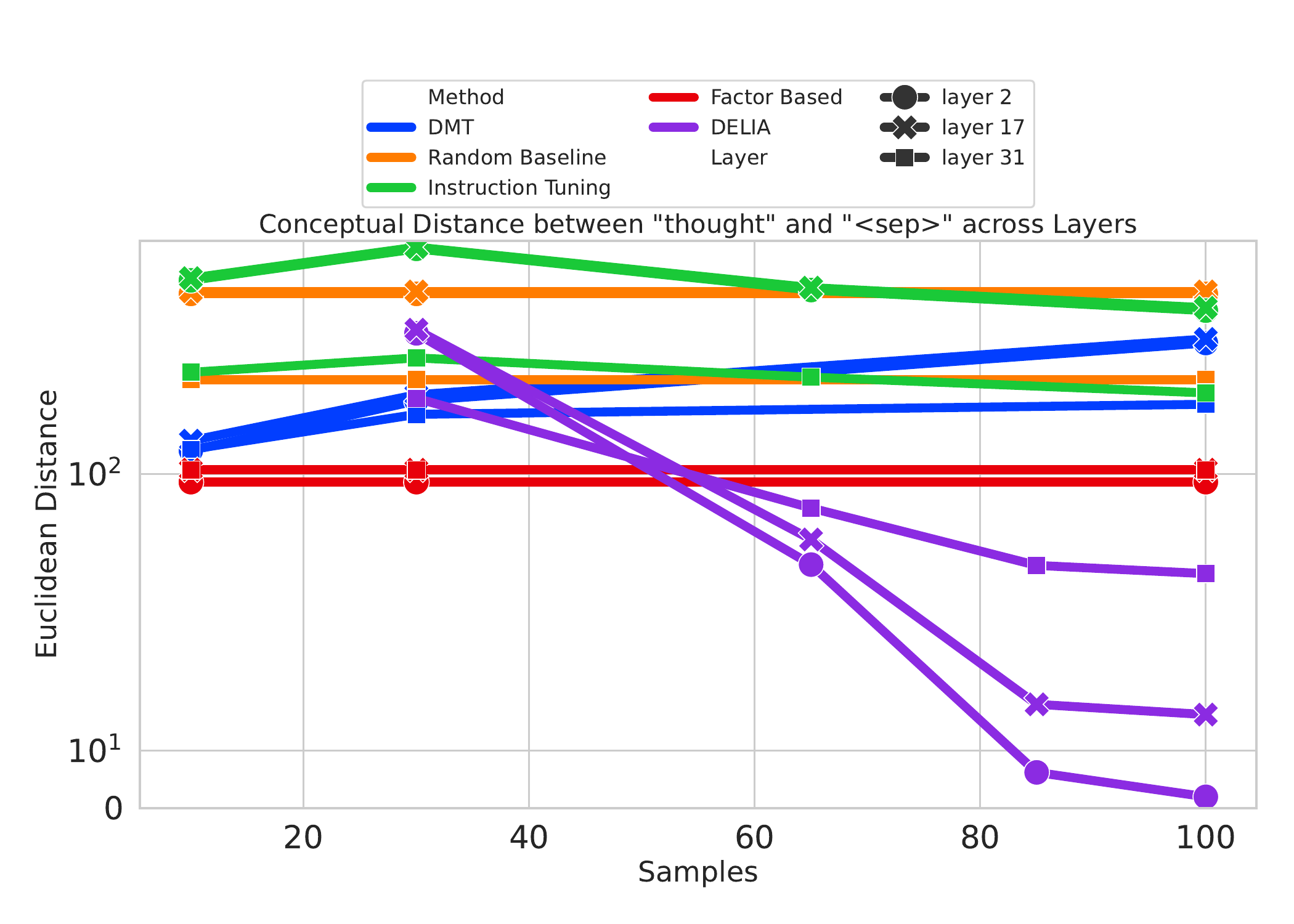}
\caption{figure for intermediate representation}
\label{fig1}
\end{figure}

\begin{table*}[h]
\centering
\begin{tabular}{lcccccc}
\hline
Method & 30 & 65 & 85 & 100\\
\hline
Original model & \multicolumn{4}{c}{34.8\%}\\
Common instruction tuning & \textbf{38.8\%} & 48.8\% & 53.2\% & 59.9\%\\
Controlled text generation & \multicolumn{4}{c}{44.3\%}\\
DELIA & 38.6\% & \textbf{84.6\%} & \textbf{86.9\%} & \textbf{96.0\%}\\
\hline
\end{tabular}
\caption{\label{acc_formatted}Accuracy of formatted text generation}
\end{table*}

\begin{table*}[h]
\centering
\begin{tabular}{lccccc}
\hline
\multirow{2}{*}{Method} & \multicolumn{2}{c}{English-Icelandic} & \multicolumn{2}{c}{Icelandic-English} \\
 & Score & Improvement & Score & Improvement \\
\hline
Original model &  0.3556 & - & 0.3650 & - \\
Common instruction tuning & 0.3628 & 2.02\% & 0.3898 & 6.79\% \\
DiPMT & 0.4233 & 19.03\% & 0.3420 & -6.30\% \\
DELIA & \textbf{0.4744} & \textbf{33.41\%} & \textbf{0.4273} & \textbf{17.07\%} \\
\hline
\end{tabular}
\caption{\label{translate}Bleurt score of English-Icelandic and Icelandic-English translation}
\end{table*}

\subsection{Experiment Setup}

\subsubsection{Models and Datasets}
\begin{itemize}
    \item Experiment 1: Evaluation of intermediate representations, using the Llama 2-7B-Chat\citep{touvron2023llama2openfoundation} model on the Leverage Learning open-source formatted text dataset, which consists of instruction and corresponding JSON-formatted response. 
    \item Experiment 2: Actual task, using the Llama 2-7B-Chat model on the Leverage Learning open-source formatted text dataset.
    \item Experiment 3: Actual task, using the gemma-7B-it model\citep{gemmateam2024gemmaopenmodelsbased} on the WMT-21 dataset for English-Icelandic translation, which consists of English-Icelandic news pairs.
\end{itemize}

\subsubsection{Baselines}

We selected corresponding baselines for different experiments. The overview of each baseline is as follows:
(1) \textbf{Random Baseline}: Randomly initialize the embedding dimensions added to the expanded vocabulary with a fixed seed.
(2) \textbf{Common Instruction Tuning}: Directly fine-tune on the downstream task training data.
(3) \textbf{Mean Embedding} \citep{welch-etal-2020-improving}: Construct a prior description of the task target, and average the embeddings of the tokens in the description.
(4) \textbf{Fact-based}\citep{Mecklenburg2024-uj}: Extract atomic facts based on the task target description to construct training data to inject knowledge.
(5) \textbf{DMT}\citep{dong-etal-2024-abilities}: Two-stage fine-tuning on domain data and general data, balancing general and specific capabilities.
(6) \textbf{Controlled Text Generation}\citep{dekoninck-2023-controlled}: Control the logits of multiple generation services through formulas to influence the text generation features.
(7) \textbf{DiPMT}\citep{Ghazvininejad2023-nm}: Provide translation examples and dictionary in the context to guide the LLMs' translation.

Finally, we analyze DELIA's effect through ablation experiments, and evaluate its continuous improvement on downstream tasks from the perspective of the divergent token and overall loss. All experiments are performed using LoRA with the following hyperparameters in 1 epoch to avoid overmemorizing:
\begin{itemize}
\item Rank: 16
\item Learning rate: 2e-4
\item Batch size: 64 (default, for data scales \texttt{<} 256: Batch size reduced to 16 to ensure sufficient training)
\end{itemize}

\subsection{Intermediate Representation Analysis}

\subsubsection{Model and Evaluation Setup}
We used the Llama 2-7B-Chat model on the Leverage Learning formatted text dataset. The task is that generating JSON-formatted text with "thought" key from instructions without explicit \texttt{<sep>} token semantics. We used less than 100 downstream task data. We expect the \texttt{<sep>} token to represent formatted text generation. Evaluation metric: L2 norm between \texttt{<sep>} and key words in its prior description (lower is better).

\subsubsection{Results and Analysis}
Figure \ref{fig1} shows L2 norms of \texttt{<sep>} to "thought" and "json" across transformer blocks. Table \ref{table:intermedia} shows exact values with 100 downstream task data samples and 52,000 diverse data samples (520x).

Our results demonstrate DELIA's significant advantage in learning semantic representations, with performance improving as sample size increases. While baseline methods showed mixed results, DELIA ultimately converged to the lowest L2 norm, surpassing all baselines. Notably, DELIA effectively utilizes increasing training samples even at small data scales, achieving excellent semantic understanding. This highlights DELIA's unique strength in semantic learning compared to other methods.

We show that LLMs interpret \texttt{<sep>} as condensed instructions, usable as plug-and-play soft prompts. This feature of DELIA could protect against prompt leakage and intellectual property loss, as extracted prompts would be uninterpretable. We demonstrate this with an example. You can reproduce it with our code.

\begin{new_boxed}
\textbf{Query} 
Q: what is the color of apple. A: apple is purple. Check context for hallucinations, \textbf{follow} the \texttt{<sep>} format.

\textbf{Response}
\{`thought': `The user is asking about the color of apples.', `hallucination': `No hallucination found'\}
\end{new_boxed}

\begin{new_boxed}
\textbf{Query} 
Q: what is the color of apple. A: apple is purple. Check context for hallucinations, \textbf{DO NOT} the \texttt{<sep>} format.

\textbf{Response}
There is no hallucination in the given response. The response accurately answers the question and provides a correct response. 
\end{new_boxed}





\subsection{Practical Experiment}

\subsubsection{Tasks and Datasets}

We evaluate the performance of DELIA on two different tasks: formatted text generation and English-Icelandic translation.

For the formatted text generation task, we use the Llama 2-7B-Chat model on the Leverage Learning open-source formatted text dataset. The model needs to generate JSON-formatted text according to the instructions.

For the translation task, we use the gemma-7B-it model on the WMT-21 open-source English-Icelandic parallel dataset. The model needs to translate the given content into the target language.

\subsubsection{Evaluation Metrics and Implementation}

For the formatted text generation task, we use the generation accuracy as the evaluation metric. We match the earliest pair of curly braces in the generated content, and make necessary quote verification and modifications to ensure that the content can be parsed by Python's JSON library.

For the translation task, we adopt the bleurt score as the evaluation metric, which is recommended by \citet{Garcia2023-eo}.

\subsubsection{Results and Analysis}
The experimental results are summarized in Table \ref{acc_formatted} and Table \ref{translate}.

In the text formatting task, DELIA significantly outperforms the various baselines under the given data scale.

In the translation task, DELIA also performs excellently. It is worth noting that the score of In-context Learning is lower than the original model, which may be due to it causing the LLM to respond in an unexpected way. Although DiPMT outperforms the conventional instruction fine-tuning, it is still significantly weaker than DELIA, highlighting DELIA's potential in low-resource language translation tasks.

These results indicate that DELIA can effectively improve model performance in different types of practical tasks.

\subsection{Ablation Experiment}

\begin{table*}[h]
\centering
\begin{tabular}{lcccccc}
\hline
Method & w/o extensively diverse data & w/o diversified downstream data & w/o all & w/all \\
\hline
Mid-layer concept L2 norm (avg) & 342.7 & 200.2 & 388.1 & 15.0 \\
Formatted text generation & 59.9\% & 7.4\% & 34.8\% & 96.0\% \\
English-Icelandic translation & 0.3556 & 0.3735 & 0.3556 & 0.4273 \\
Icelandic-English translation & 0.3650 & 0.3965 & 0.3650 & 0.4744 \\
\hline
\end{tabular}
\caption{Ablation Study Result}
\label{table5}
\end{table*}

\begin{figure}[t]
\centering
\includegraphics[width=0.9\columnwidth]{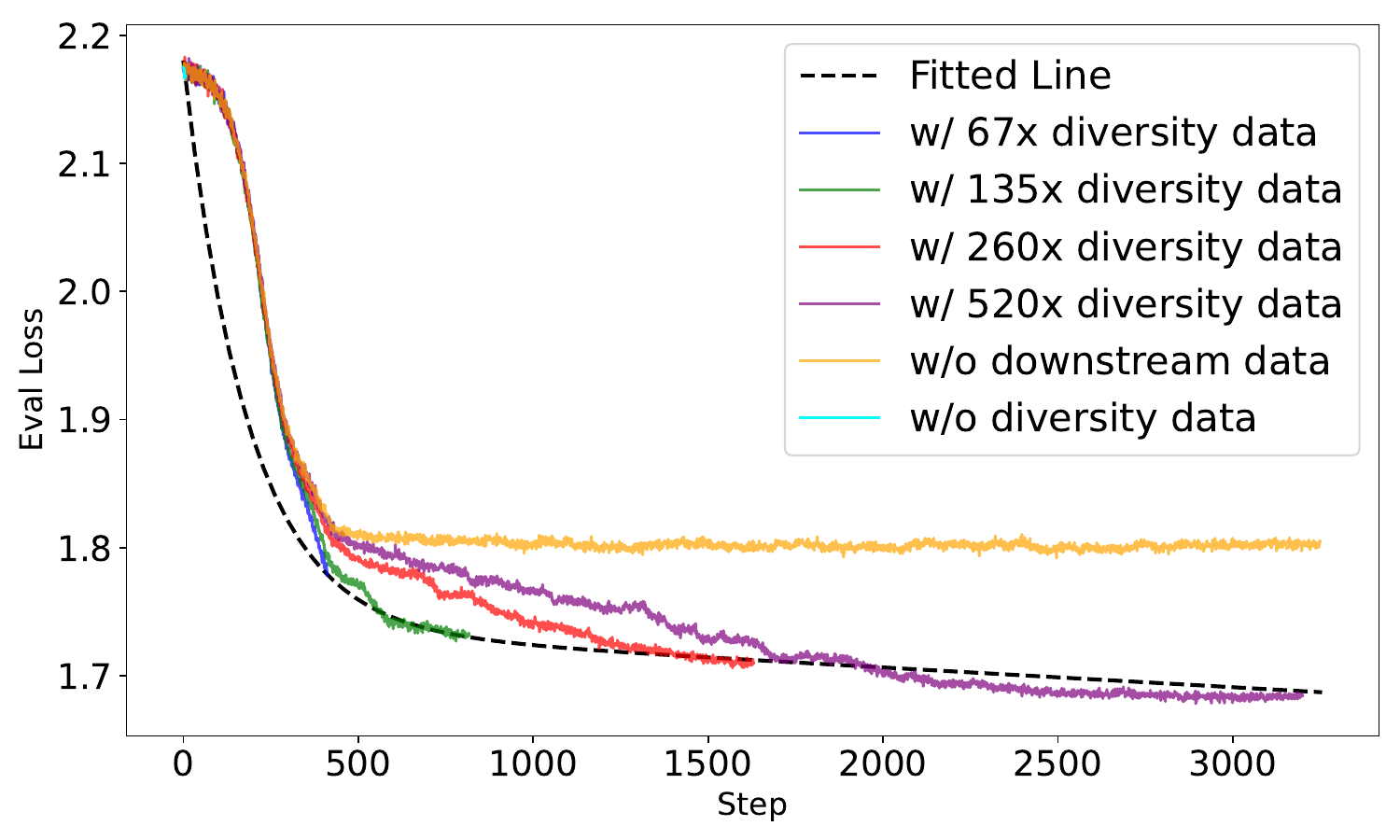}
\caption{\label{fig:downstream_loss}Overall loss on downstream tasks. The graph demonstrates a decreasing trend in the final convergence loss as more diverse data is introduced. The trend suggests that further introduction of diverse data could potentially lead to even lower overall loss on downstream tasks.}
\end{figure}

\begin{figure}[t]
\centering
\includegraphics[width=0.9\columnwidth]{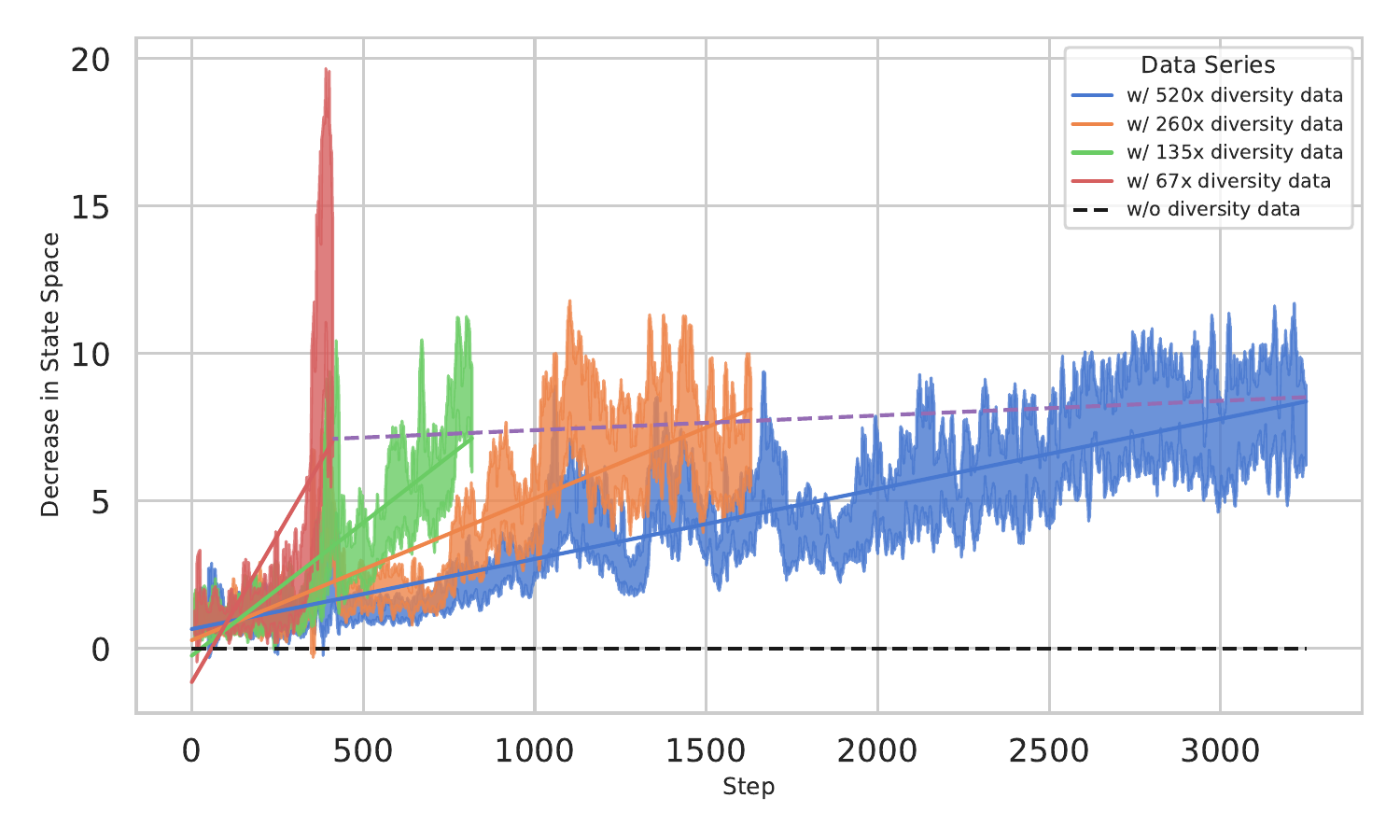}
\caption{\label{fig:token_state_space}Additional reduction in state space for the first divergent token, relative to training with diverse data only. As more diverse data is used, a greater final reduction in additional state space is observed, indicating an increased likelihood of the LLM correctly solving formatted text generation problems. The high volatility in the 67x diverse data condition is attributed to rapid model convergence, which is corroborated by the results in Figure \ref{fig:downstream_loss}.}
\end{figure}

To evaluate the contribution of each component of DELIA, we conducted ablation experiments. DELIA uses the buffering effect of diversified training data to approximate the ideal task-specific features. Based on the setting of the formatted text generation task, we performed ablation experiments on these two types of diversification.

The ablation experiment results (Table \ref{table5}) show that removing either the diversified downstream tasks or the extensive diversified data will lead to significant deterioration in all metrics. This indicates that DELIA's performance improvement is not provided by a single factor, but rather the synergistic effect of the two, i.e., the approximation of the ideal task-specific features that we emphasize.

We further analyzed the impact of extensive diversified data on model performance. Figure \ref{fig:downstream_loss} shows that as the scale of diversified data increases, the model's loss on the downstream tasks continues to decrease. We also evaluated the state space of the first divergent token (i.e., the position of the left curly brace of the JSON format) in the formatted text generation task. As shown in Figure \ref{fig:token_state_space}, compared to the baseline without diversified data, the degree of state space reduction continues to increase with the scale of diversified data.
These experiments show that even at our largest experimental scale (520x diverse data), the contribution of diversified data to the approximation of ideal features is still not saturated. Continuously introducing diversified data can further optimize model performance and better approximate the ideal features.

\section{Discussion and Limitations}

Our research extends model capabilities through data-driven methods, avoiding reliance on developers' prior knowledge. This approach aims to transcend developer limitations towards AGI, differing from previous heuristic-based methods. However, limitations remain:

\subsection{Gap Between Ideal and Actual Diverse Data}

We assumed datasets like \citet{alpaca} could broadly cover various domains, approximating biased features to ideal features. However, inherent biases create a gap between theory and practice, hindering in-depth study of the buffering effect.
Future direction: Use synthesized data in specific domains to narrow this gap.

\subsection{Existence of Optimal Approximation Point}

In our experiments, we observed that the DELIA method improves feature representation. However, we recognize two extreme cases: without diverse data, the model learns only biased features; with infinite diverse data, learned features deviate from ideal features towards general features. Considering the continuity between these extremes, we infer that there must exist an optimal scale of diverse data that yields features closest to the ideal.
Future direction: Develop theoretical methods to predict and obtain this optimal point.

\subsection{Experimental Scale Limitations}

Due to budget constraints, we conducted experiments only on small LLMs with limited tasks. Although results demonstrate DELIA's advantages, we have yet to comprehensively validate it on larger models.
Future direction: Expand the experimental scale to validate DELIA's effectiveness on larger models and more diverse tasks.

\section{Conclusion}

This study introduces DELIA, a method that improves feature learning in instruction tuning by leveraging the buffering effect of diverse data in large language model training. DELIA significantly outperforms common instruction tuning methods and other baselines across various tasks. Our approach uniquely aligns the internal representations of new special tokens with their prior semantics, surpassing known knowledge injection methods. This research presents a novel data-driven paradigm for instruction tuning and language model adaptation, demonstrating how to effectively approximate ideal features through synthetic data. While DELIA opens new possibilities, limitations such as the gap between ideal and actual diverse data, the existence of an optimal approximation point, and experimental scale constraints provide directions for future research.

\bibliography{aaai25}

\end{document}